# Ranking Under Uncertainty

Or Zuk[1,*]　　　　　Liat Ein-Dor[1,2,*]　　　　　Eytan Domany[1]

[1] Dept. Phys. Comp. Systems
Weizmann Inst. of Science
Rehovot, 76100, Israel
{*or.zuk/eytan.domany*}*@weizmann.ac.il*

[2] Machine Learning Group
Intel Research Labs
Haifa, 31015, Israel
*liat.ein.dor@intel.com*

## Abstract

Ranking objects is a simple and natural procedure for organizing data. It is often performed by assigning a quality score to each object according to its relevance to the problem at hand. Ranking is widely used for object selection, when resources are limited and it is necessary to select a subset of most relevant objects for further processing. In real world situations, the object's scores are often calculated from noisy measurements, casting doubt on the ranking reliability. We introduce an analytical method for assessing the influence of noise levels on the ranking reliability. We use two similarity measures for reliability evaluation, Top-K-List overlap and Kendall's $\tau$ measure, and show that the former is much more sensitive to noise than the latter. We apply our method to gene selection in a series of microarray experiments of several cancer types. The results indicate that the reliability of the lists obtained from these experiments is very poor, and that experiment sizes which are necessary for attaining reasonably stable Top-K-Lists are much larger than those currently available. Simulations support our analytical results.

## 1 Introduction

Ranking objects by their importance, quality or other properties of interest is a natural procedure taking place in a wide variety of fields. Web pages are ranked according to their relevance to a given query, tennis players are ranked by their achievements, scientific journals are ranked by their impact etc. This diversity of applications stems from the fact that ranking is a simple and straightforward way for organizing data in an informative manner, which helps to manipulate the data more efficiently, unravel internal relations between objects, and create a global picture regarding their relevance to the considered problem.

In an ideal noiseless world, one can associate a pure and objective quality or score to each given object. Sorting these numbers yields the 'true' ranking of the objects. These numbers are sometimes imposed naturally by the problem, (e.g. ranking stocks by their growth rate), but also appear in situations where no precise values are given, for example they maybe the output of some rank aggregation algorithm, where each individual ranking represents a biased view of the world [1, 2]. However, in many real world scenarios, only a noisy version of the object's values is available, while their true values are unknown. This noise may distort the observed ranking (which is based on the noisy values) with respect to the true one (which is based on the correct, noise-free value associated with each object). In this paper we explore the influence of the noise level on the reliability of the observed ranking for a large number of objects. For this purpose we develop an analytical framework which gives, for a given noise level, the probability distribution of the discrepancy between the observed and true rankings. This framework enables us to give confidence intervals for the reliability, thus answering questions such as "what noise level allows one to produce a ranking whose discrepancy with the true one is less than $\epsilon$ with probability larger than $1 - \delta$"?

One may wonder why are we concerned with the similarity of the *rankings* on the first place, when we have access to the more informative numerical values (usually ranks are used when no such numerical values are available, for example when we want to rank objects based on partial data from various sources). Our answer is that typically ranking of objects is important in situations of *limited resources*, where objects with different (measured) ranks may be treated differently.

---
*[*] These authors contributed equally to this work.

In this case, the rank itself and not the numerical value is the important quantity - for example a tiny difference of a few milliseconds in the 100m olympics final might separate between the runner ranked first (the gold medalist), and the runner-up. Here the 'resource' is, say, the first prize. Another example for the role of ranking in a limited resources environment is of a search engine which ranks web pages based on some complicated 'relevance' score, indicating the match of each page to the submitted query. The rank of a page has a great influence on the probability that a typical user will access it, as the bounded resource is the time and efforts the user is willing to spend on finding the desired page. Similarly, university departments may rank their candidates based on scores they have obtained in some tests, and since the number of available positions is bounded, the rank of a specific candidate is what determines whether he will be admitted to the department or not.

Due to its simplicity and efficiency, ranking serves as a principal or auxiliary selection mechanism in many feature selection algorithms [3, 4]. The main objective of variable ranking is not necessarily building classifiers. A more common application of this mechanism is finding the most relevant variables to the considered problem. This type of application is highly useful in problems from various areas, e.g. gene selection in microarray analysis [5–11], where the aim is to find a set of potential drugs by finding the set of genes with the highest discriminative power between two (or more) patient populations. Commonly used discriminative power measures are for example Pearson correlation and mutual information. A set of relevant variables is obtained by selecting the $K$ variables with the highest scores, where $K$ is determined either based on resources constraints, or, for classification problems, using cross validation methods. Ranking individual features is optimal for Naive-Bayes classifiers. For other classifiers, it may still give good performance, but sometimes it is needed to consider the ranking of combinations of features.

As was mentioned before, in real world situations we do not have access to the true object score, which we define, in the context of classification, as the score that would have been obtained using *all* possible labeled examples; the measured value of the score is based on information derived from the (relatively small number of) available labeled samples. This deviation of the measured score from the true one can be regarded as noise, which can potentially change the feature ranking, and thus also the composition of the top scoring list. Thus the noise in variable ranking problems is due to the effect of *finite sampling* from an entire population, and our methods can be used to estimate the number of samples necessary to identify the most relevant features in a classification problem of high dimensional data. We demonstrate the accuracy of our results for microarray datasets from several cancers. We show that the number of samples necessary in order to get a reliable approximation of the true ranking may be surprisingly high.

The main concern of this paper is the stability of the ranking - i.e. how similar is the ranking that was carried out in the presence of noise to the 'true' ranking, based on the uncorrupted scores? In section 2, we formulate the mathematical model we propose in order to answer this question. We note that one can measure the agreement between two rankings in different ways. We choose two permutation similarity measures to evaluate this agreement, each highlighting a different aspect of the ranking, and explain the importance of each of them. In section 3, we describe the solution of our model - i.e. we introduce a mathematical expression for the distribution of the agreement between the measured and true rankings, and compute (or approximate) its first two moments. This is carried out for the two chosen measures, in the limit where $N$, the number of ranked objects, is large. We compare our results to simulated data (sec. 4) and to real-world microarray data (sec. 5), and find good fit with our analytical calculations. The last section is devoted to discussion and future directions.

## 2　Problem Representation

Assume $N$ objects, each with an associated real number, $\vec{r} = (r_1, .., r_N)$ representing the 'true' value of some property of the object (i.e. the 'score'). If these numbers are given, one can obviously rank the objects based on their values, to get the true ranking $\pi = \big(\pi(1), .., \pi(N)\big)$. Suppose now that we do not have access to the true values $r_i$, but to some noisy version of them $s_i = r_i + Z_i$ where the $Z_i$ are i.i.d. centered random variables drawn from a probability distribution $G \equiv G(\sigma)$ with variance $\sigma^2$. We argue that in many applications the measurement noise is Gaussian or close to Gaussian, as often the noise is due to accumulation of many random events, and we will therefore assume that $G = N(0, \sigma^2)$ from now on for simplicity. Still, our analytic derivations can be modified and performed for other families of noise distributions. The measured values $s_i$ are possibly different from the true values $r_i$, and may thus induce a different ordering of the $N$ objects, denoted $\pi_\sigma$. We are interested in the behavior of the similarity $c(\pi, \pi_\sigma)$, as a function of the noise level $\sigma$. Here $c \equiv c(\pi_1, \pi_2)$ is any similarity measure for permutations, as will be discussed below.

## 2.1 Permutation Similarity Measures

Due to measurement noise, the measured ranking is typically not identical to the true ranking, and can be viewed as only an approximation of it. In order to assess how good is the approximation, one needs a quantitative measure for comparing permutations. Distance and similarity measures for permutations have a rich literature in various fields - see [12] for a survey. The specific choice should be application dependent. As a matter of convention, we will use in this work *similarity* measures (one can easily obtain a similarity measure from a given metric). A straightforward choice is to treat the rankings simply as vectors of natural numbers in $\{1, .., N\}$, and use Spearman's rank correlation coefficient. Another approach, popular in genome-rearrangement problems in computational biology, seeks the minimal number of 'atomic operations' (e.g. transpositions) one needs to perform on $\pi_1$ in order to reach $\pi_2$ (for transpositions this is called the Cayley distance, see [13]). We chose to focus on the two following similarity measures:

1. Kendall's $\tau$ rank correlation [14] - This measure simply counts the number of pairs whose relative ordering in the two permutations agrees. Define $\tau_{ij}(\pi_1, \pi_2)$ to be 1 if $\pi_1$ and $\pi_2$ agree on the order of elements $i$ and $j$, i.e. $\tau_{ij} = \Theta\big((\pi_1(i) - \pi_1(j))(\pi_2(i) - \pi_2(j))\big)$, where $\Theta$ is the Heaviside function. Then Kendall's $\tau$ is given as[1]:

$$\tau(\pi_1, \pi_2) = \frac{1}{\binom{N}{2}} \sum_{i<j} \tau_{ij}(\pi_1, \pi_2). \quad (1)$$

Unlike Spearman's rank-correlation, Kendall's $\tau$ has a direct intuitive meaning - it represents the probability that the two rankings agree on the ordering of a given pair of objects. It has became widely used in non-parametric statistics.

2. *K-top* overlap measure - In many cases the relevant information is not the entire ranking but the list of $K$ objects that are ranked highest. This is the typical case in limited resources scenarios, where the number of objects is much higher than what available resources can handle, and a subset of $K$ top ranked objects needs to be selected. For such events, a natural measure of similarity is the overlap $f$ between two *Top-K-Lists* (TKLs):

---

[1] We have chosen to use this definition throughout the paper and not the standard Kendall's $\tau$ whose range is $[-1, 1]$, in order to preserve the meaning of this similarity measure: the fraction of pairs of elements whose ordering agree on both rankings.

$$f(\pi_1, \pi_2) = \frac{1}{K} \sum_{i=1}^{N} \Theta\big(K+1-\pi_1(i)\big)\Theta\big(K+1-\pi_2(i)\big). \quad (2)$$

The two proposed measures capture different aspects of the ranking. Kendall's $\tau$ gives an overall 'smooth' agreement measure, whereas the TKL overlap introduces a sharp cutoff and is not influenced by the relative ranking above or below the cutoff. We have chosen these two simple measures to demonstrate our analytical approach. More complicated measures for comparing two rankings or TKLs are sometimes needed (for example, only the top few dozens of a search engine's results are usually looked at, where pages ranking higher within the first dozens are are more likely to be accessed than pages ranking lower), see for example [15, 16]. Nevertheless, many of these measures can be represented as a combination of the two similarities we have introduced, and therefore an appropriate modification of our techniques is applicable also for them. By our definitions, both $f$ and $\tau$ range between 0 to 1, where higher values indicate higher similarity between the permutations, yet the actual values of $f$ and $\tau$ are not comparable in a meaningful way.

## 3 Analytical Solution

Recall that $\vec{s} = \vec{r} + \vec{Z}$ is the noisy version of $\vec{r}$. For simplicity we assume that the distributions of the noise variables $Z_i$ are identical independent zero-mean Gaussians, $Z_i \sim N(0, \sigma^2)$. Our aim is to explore the influence of the noise level, $\sigma^2$, on the reliability of the measured ranking $\pi_n$. We define *reliability* as the probability that the similarity between the measured and the true permutation exceeds a certain threshold, $1 - \epsilon$, for a given noise level. If we require that the reliability exceeds the value of $1 - \delta$, we have obtained a confidence interval for the similarity, and achieved *Probably Approximately Correct* [17] (PAC) ranking of the $N$ objects. We use the two aforementioned quantities $\tau$ and $f$ to measure similarity between permutations. We present an analytical evaluation of reliability in the large $N$ regime, by calculating the probability distribution of the similarity functions $\tau$ and $f$. Clearly, both overlap measures depend on the precise values of the scores vector $\vec{r}$. For large $N$, one can approximate these values by some probability density function $q$, whose support is contained in some (possibly infinite) interval $[a, b]$. This approximation is at the heart of our analytical solution. Our intention is to express the distribution of overlap measures $\tau$ and $f$ in terms of the distribution $q$, and the noise level $\sigma$. As is shown next, for large $N$ both measures can be represented as a sum of weakly correlated indicator random variables, which suggests that their distribution can be well approximated by a Gaussian. This observation leaves us only with the computation of the first two moments in order to get the (approximate) probability distribution. Although we have not proved a Central-Limit-Theorem for $\tau$'s and $f$'s distributions, simulations results presented in section 4 justify our Gaussian approximation.

### 3.1 Kendall's rank-correlation

In this section we compute the first two moments of Kendall's $\tau$, as a function of the noise level $\sigma$. Let $r_i < r_j$ be two original values, and let $s_i, s_j$ be their noisy measurements. The difference $(s_i - r_i) - (s_j - r_j)$ has a Gaussian distribution with st.d. $\sqrt{2}\sigma$. This gives:

$$Pr(\tau_{ij} = 1) = Pr(s_i \leq s_j) = \Phi\left(\frac{r_j - r_i}{\sqrt{2}\sigma}\right), \quad (3)$$

where $\Phi(X)$ is the standard Gaussian cumulative distribution. Since $r_i, r_j$ are i.i.d. drawn from the distribution $q$, the $\tau_{ij}$ are identically distributed (though they are not independent!), and by linearity of expectation, we get:

$$\mu_\tau = E(\tau_{12}) = \iint_{-\infty}^{\infty} q(r_1)q(r_2) \Phi\left(\frac{|r_2 - r_1|}{\sqrt{2}\sigma}\right) dr_1 dr_2. \quad (4)$$

The above integral may be either solved numerically, or, for some specific distributions $q$, analytically. For example, if $q$ itself is a Gaussian with st.d. $\sigma_q$, then eq. 4 becomes:

$$\mu_\tau = \frac{1}{2} + \frac{tan^{-1}(\sigma_q/\sigma)}{\pi}. \quad (5)$$

When computing the second moment of $\tau$ one has to account for the correlations between the $\tau_{ij}$'s. The computation is simplified by noticing that $\tau_{ij}$ and $\tau_{kl}$ are uncorrelated, whenever $i, j, k, l$ are all different:

$$\Sigma_\tau^2 = E(\tau^2) - \mu_\tau^2 = E\left[\frac{1}{\binom{N}{2}}\left(\sum_{i<j}\tau_{ij}\right)^2\right] - \mu_\tau^2 =$$

$$\frac{1}{\binom{N}{2}}\left[E(\tau_{12}^2) + 2(N-2)E(\tau_{12}\tau_{13}) + \binom{N-2}{2}E(\tau_{12}\tau_{34})\right]$$

$$-\mu_\tau^2 = \frac{1}{\binom{N}{2}}\left[\mu_\tau + (3-2N)\mu_\tau^2 + 2(N-2)E(\tau_{12}\tau_{23})\right]. \quad (6)$$

In order to evaluate $E(\tau_{12}\tau_{23})$ we need to consider different orderings of $r_1$, $r_2$ and $r_3$ separately (and exploiting symmetries in the orderings):

$$E(\tau_{12}\tau_{23}) = \iiint_{-\infty}^{\infty} q(r_1)q(r_2)q(r_3) dr_1 dr_2 dr_3$$

$$\left\{2 \iiint_{r_1 < r_2 < r_3} \Theta(s_2 - s_1)\Theta(s_3 - s_2) DZ_1 DZ_2 DZ_3 + 4 \iiint_{r_2 < r_1 < r_3} \Theta(s_1 - s_2)\Theta(s_3 - s_1) DZ_1 DZ_2 DZ_3\right\}, \quad (7)$$

where $DZ = dZ\, e^{-Z^2/2\sigma^2}/\sqrt{2\pi}\sigma$ and $s_i = r_i + Z_i$.

The above integral can be generally solved numerically, and plugged into eq. 6, to get $\Sigma_\tau$. We approximate the distribution of $\tau$ by a Gaussian, $\tau \sim N(\mu_\tau, \Sigma_\tau^2)$. This approximation can be used to obtain *PAC*-like confidence bounds on the value of $\tau$, for given $\epsilon$ and $\delta$.

### 3.2 Top-K-List overlap

In this section we use $f(\pi_1, \pi_2)$ as the similarity measure (denoted hereafter as $f$). We denote by $\alpha = K/N$ the fraction of objects included in the TKL, and calculate $P_{\sigma,\alpha}(f)$, the probability of obtaining an overlap $f$ between the observed and the true TKLs. For simplicity we index the objects according to their true rankings (this does not imply or necessitate knowledge of the true rankings); hence $i = 1...K$ are the indices of the objects that compose the true TKL. An object $j$ will be selected by our ranking if $s_j \geq x$ where $x$ is the threshold that separates the top $K$ objects from the rest. The threshold $x$ itself is a random variable, which can be defined for example as the $(1 - \alpha)$-quantile of the $s_j$'s, and is distributed according to some probability density function $u(x)$. The probability of an object *not being selected* is

$$P(x, r, \sigma) = \int_{-\infty}^{x} \frac{1}{\sqrt{2\pi}\sigma} \exp\left(-\frac{(s - r)^2}{2\sigma^2}\right) ds. \quad (8)$$

We can now express $P_{\sigma,\alpha}(f)$ by summing over all possible assignments of binary variables $\vec{h} = (h_1, \ldots, h_N)$, which indicate whether object $j$ was included in the TKL ($h_j = 1$) or not ($h_j = 0$), and by integrating over all possible values of the threshold $x$. We have

$$P_{\sigma,\alpha}(f) = \frac{G_{\sigma,\alpha}(f)}{\int_0^1 df G_{\sigma,\alpha}(f)}, \quad (9)$$

where the denominator is a normalization factor and

$$G_{\sigma,\alpha}(f) \equiv$$

$$\int_{-\infty}^{\infty} dx \sum_{h \in \{0,1\}^N} \left\{\delta\left(\sum_{j=1}^{N} h_j - K\right) \delta\left(\sum_{j=1}^{K} h_j - fK\right)\right.$$

$$\left.\prod_{j=1}^{N}\left[(1 - h_j)P(x, r_j, \sigma) + h_j(1 - P(x, r_j, \sigma))\right]\right\}. \quad (10)$$

The first Kronecker delta function ensures that the total number of selected objects is $K$, and the second

- that $fK$ of these belong to the true TKL. The sum is over all binary vectors $\vec{h}$ of size $N$ and the product over all objects is the probability of each $\vec{h}$. We implicitly assumed that each value of the threshold has the same a-priori probability[2]. Using the integral representation of the two Kronecker's delta functions and exchanging the order of the summation and product, we sum over the $h_j$'s in eq. 10 to obtain:

$$G_{\sigma,\alpha}(f) = \int_{-\infty}^{\infty} dx \int_{-\pi}^{\pi} \frac{dydz}{(2\pi)^2} \times$$

$$\exp\left(iyN(1-\alpha) + izN\alpha(1-f)\right) \times$$

$$\exp\left(\sum_{j=1}^{K} \ln(1 + P(x,r_j,\sigma)(e^{-i(y+z)} - 1))\right) \times$$

$$\exp\left(\sum_{j=K+1}^{N} \ln(1 + P(x,r_j,\sigma)(e^{-iy} - 1))\right). \quad (11)$$

Since $N >> 1$ the sums over the objects can be replaced with integrals of their probability distribution to give:

$$G_{\sigma,\alpha}(f) \sim \int_{-\infty}^{\infty} dx \int_{-\pi}^{\pi} \frac{dydz}{(2\pi)^2}$$

$$\exp\left(-N\left[-iy(1-\alpha) - iz\alpha(1-f)\right]\right) \times$$

$$\exp\left(-N\left[-g(b,r_\alpha,x,y,z) - g(r_\alpha,a,x,y,0)\right]\right), \quad (12)$$

where $[a,b]$ is some interval containing the support of $q$, $g(x_1,x_2,x_3,y,z) \equiv \int_{x_2}^{x_1} q(r) \ln(1 + P(x_3,r,\sigma)(e^{-i(y+z)} - 1))dr$, and $r_\alpha$ is the $1-\alpha$ quantile of $r$, i.e. $\int_{r_\alpha}^{b} q(r)dr = \alpha$. $r_\alpha$ is the threshold value of the true TKL, namely an object is in the true list iff the true value of its score exceeds $r_\alpha$. At this stage one can apply the saddle point method [18] to solve the multiple integral in eq. 12. The saddle point equations are obtained by setting to zero the partial derivatives of the function

$$F(x,y,z,f) \equiv -iy(1-\alpha) - iz\alpha(1-f) - \quad (13)$$
$$g(b,r_\alpha,x,y,z) - g(r_\alpha,a,x,y,0)$$

with respect to $x, y, z$. Here we denoted by $P_x(x, r, \sigma)$ the partial derivative of $P(x, r, \sigma)$ with respect to $x$. The saddle point is thus given as the solution of the following system of three algebraic equations:

$$\alpha(1-f) = \int_{r_\alpha}^{b} \frac{P(x,r,\sigma)e^{-i(y+z)}q(r)}{1 + P(x,r,\sigma)(e^{-i(y+z)} - 1)}dr,$$

$$(1-\alpha) - \alpha(1-f) = \int_{a}^{r_\alpha} \frac{P(x,r,\sigma)e^{-iy}q(r)}{1 + P(x,r,\sigma)(e^{-iy} - 1)}dr,$$

---

[2]The precise form of the threshold density function $u(x)$ has a negligible effect on the results in the large $N$ limit.

$$\int_{r_\alpha}^{b} \frac{P_x(x,r,\sigma)(e^{-i(y+z)} - 1)q(r)}{1 + P(x,r,\sigma)(e^{-i(y+z)} - 1)}dr +$$

$$\int_{a}^{r_\alpha} \frac{P_x(x,r,\sigma)(e^{-iy} - 1)q(r)}{1 + P(x,r,\sigma)(e^{-iy} - 1)}dr = 0. \quad (14)$$

For the normalization factor (the denominator in eq. 9) the saddle point equations are those appearing in eq. 14 and an additional equation emerging from the maximization of $F(x,y,z,f)$ over $f$ yielding $z = 0$. Inserting the saddle point solutions of the numerator and denominator in eq. 9, and keeping the leading order, one obtains the following expression for $P_{\sigma,\alpha}(f)$

$$P_{\sigma,\alpha}(f) \sim \sqrt{\frac{N|R|}{2\pi|H|}} e^{-N\left[F(f) - F(f_0)\right]}, \quad (15)$$

where $|H|$ and $|R|$ are the determinants of the second derivative matrices $H, R$, of $F$ with respect to $(x, y, z)$ and $(x, y, z, f)$ respectively. $f_0$ is the value of $f$ that minimizes $F$; it is determined by $\alpha, \sigma$ and the distribution $q$. Note that $P_{\sigma,\alpha}(f)$ has a sharp maximum at $f_0$, and it approaches $\delta(f - f_0)$ as $N \to \infty$. This means that in this limit the overlap between the measured and the true TKLs is not influenced by the specific realization of the $Z_i$ r.v.s but only by their variance $\sigma^2$. For finite but large $N$, eq. 15 can be further simplified by performing a Taylor expansion of $F(f)$ around $f_0$ [19]. The first order contribution stems from the quadratic order of the expansion which results in the following Gaussian approximation:

$$P_{\sigma,\alpha}(f) \sim \frac{1}{\sqrt{2\pi}\Sigma_f} e^{-\frac{(f-f_0)^2}{2\Sigma_f^2}}, \quad (16)$$

i.e. $f \sim N(f_0, \Sigma_f^2)$ with the variance $\Sigma_f^2 = \frac{|H|}{N|R|}$.

### 3.2.1 Comparison of the two measures

The two solutions presented above, show a qualitative difference between the two measures. This difference can be expected, as the sharp cut-off in the TKL measure makes the TKL overlap much more sensitive to noise than Kendall's $\tau$, which is a 'smoother' measure. The difference is illustrated in Fig. 1, for Gaussian $q$. We have also performed asymptotic analysis (not shown) of the two similarity measures in the regime of low 'Signal-to-Noise-Ratio' ($\sigma \to \infty$), which give different first-order behaviors confirming the above observations.

## 4 Simulations

To check the accuracy of our method, and to assess the applicability of the large-$N$ expansion on which our results are based, we performed simulations in the following way:

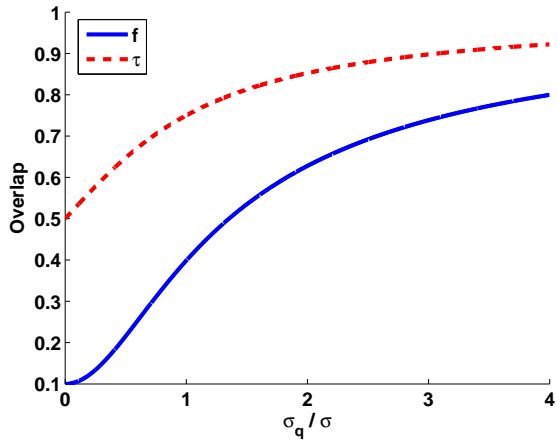

Figure 1: The expected similarity between the true and observed ranking, as a function of the signal-to-noise ratio $\sigma_q/\sigma$, for Gaussian $q$. The TKL overlap $f$ (shown here for $\alpha = 0.1$) can be much lower than Kendall's $\tau$, especially for intermediate noise levels. Similar behavior was observed for other values of $\alpha$.

---

**Simulate Ranking**

Input : $q(r)$, $N$, $K$, $\sigma_{min}$, $\Delta\sigma$, $\sigma_{MAX}$ and $N_I$
1. Draw $N$ true i.i.d. object values $(r_1, ..r_N)$ from the distribution $q(r)$, and rank them in a decreasing order to obtain the true ranking $\pi$.
2. Set $\sigma^2 = \sigma_{min}^2$.
3. Generate the observed $(s_1, ..s_N)$ by submitting the true $r_i$'s to i.i.d. additive Gaussian noise, i.e. $s_i = r_i + Z_i$, with $Z_i \sim N(0, \sigma^2)$, and sort the resulting $s_i$'s to obtain the observed ranking $\pi_\sigma$.
4. Calculate $\tau(\pi, \pi_\sigma)$ and $f(\pi, \pi_\sigma)$.
5. Repeat step (3-4) $N_I$ times. Compute the mean and variance of the recorded $\tau$ and $f$ values.
6. If $\sigma^2 < \sigma_{MAX}^2$, set $\sigma^2 = \sigma^2 + \Delta\sigma$ and go to 3.
7. Output the mean and variance of $\tau$ and $f$ as a function of $\sigma$.

---

A good match was found between simulations and analytical results as shown in the next section.

## 5 An Application to Gene Selection Using Microarrays

Personalized therapy is one of the main challenges of cancer research, requiring identification of the metastatic potential of the tumor at an early stage of the disease. The novel technology of DNA-microarrays has been recruited to perform this task, aiming at predicting the outcome of cancer patients using the gene expression profile of the primary tumor. Many microarray experiments have been carried out in an attempt to find a list of genes whose expression profile is indicative of outcome. These lists are used for two main purposes: gaining a short list of biologically relevant and potential target genes, and building classifiers discriminating between good and poor prognosis patients. Usually an outcome predictive gene list (OPGL) is derived by ranking the genes according to their relation to survival, which can be measured in several ways such as absolute correlation of the gene's expression with outcome [5, 20], mutual information between expression and outcome or other scores. An OPGL is composed of the $K$ top-ranked genes, where $K$ is determined using cross validation methods. However, the aforementioned scores suffer from uncertainty due to noise in gene expression measurements and, in particular, to the limited number $n$ of available samples from patients. Consequently, the OPGLs exhibit considerable instability under random selection of training sets, even when all samples are from the same type of cancer and the expression values are measured using the same platform [21–23]. We use our analytical framework to investigate the reliability of OPGLs obtained from several cancer studies. For gene $i$, $(i = 1,..,N)$ we denote by $c_i(n)$ the observed Pearson correlation between its expression value, as measured by the microarray, and the outcome variable, indicating the patient's survival status [5]. We denote by $c_i$ the 'true' (unknown) correlation value, which would have been obtained had we measured the correlation over the entire population of cancer patients. The 'true' score of gene $i$ is defined as the absolute Fisher $Z$ score [24, 25], $r_i = |tanh^{-1}(c_i)|$, and similarly, the observed score is $s_i \equiv s_i(n) = |tanh^{-1}(c_i)|$. The absolute value is taken since we want to treat genes with positive and negative correlation equally. Performing the Fisher transform preserves the ranking and yields two benefits: the observed Fisher scores have an approximate Gaussian distribution around their true values, and, furthermore, the variance of this distribution is not dependent on the true correlation, and is given by $\sigma^2 = 1/(n-3)$ [24, 25]. These two assertions make the Fisher-transformed scores perfectly appropriate for analysis using our mathematical model presented in Sec. 2. (Note that both assertions fail dramatically for the untransformed observed correlations).

We do not have access to the true distribution $q = q(r)$, but rather to the distribution of the $s_i$ values denoted $q'$, which is the convolution of $q$ with $N(0, \sigma^2)$ Gaussian noise. However, in all the cases we have observed, $q'$ was very well fitted by a Gaussian distribution with some variance $V_o$, indicating that the distribution of the true scores, $q(r)$, is also a Gaussian. $\sigma_q^2$, the variance of $q$, can be easily extracted from the relation $\sigma_q^2 = V_o - \sigma^2 = V_o - 1/(n-3)$. Using $q$ and $\sigma^2$ we can calculate, for given overlap and confidence requirements which we denote by $\epsilon$ and $\delta$, the number of samples necessary to obtain, with probability higher than $1 - \delta$, an OPGL whose overlap with the true OPGL exceeds $1 - \epsilon$. Application of our method to OPGL's from six different cancers is shown in Fig. 2 for the value $\alpha = K/M = 0.01$ used by several studies. We set $N$ to be the number of genes passing a preprocessing filter, thus $N$ and $K = \alpha N$ vary from one dataset to another. Our results show that the number of samples (dozens to hundreds) which were used to obtain the currently available OPGL's is much smaller than the number which is necessary to guarantee reasonable overlap with the true lists (thousands). Fig. 3 presents nice agreement between simulation with $N_I = 500$ iterations and analytical results for three out of the six datasets. Excellent match was observed for all cancers, yet only three are shown here due to space limitations.

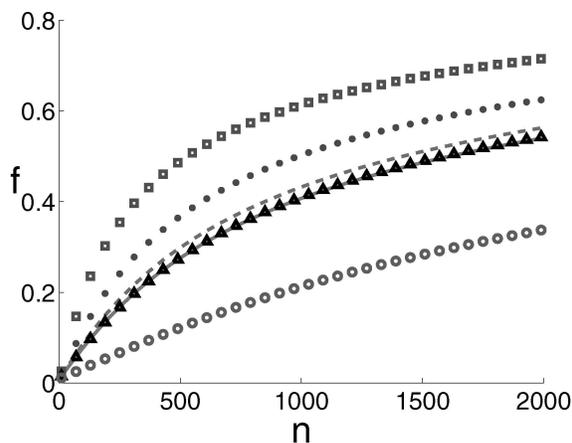

Figure 2: The expected overlap between the true and observed OPGLs is presented for microarray datasets of six cancers, as a function of the number of samples $n$, which were used to obtain the observed OPGLs. Results are displayed for Hepatocellular Carcinoma [7] (blue squares), Breast Cancer [20] (dark green dots), Breast Cancer [10] (light green dashed), Lung Cancer [6] (dark triangles), Acute Lymphocytic Leukaemia [8] (magenta solid) and pediatric ALL [9] (red circles). For very small $n$, the noise is very large and for all observed OPGLs the overlap with the true list is $\alpha$ (these experiments had $\alpha \approx 0.01$), which is the expected overlap of a randomly selected set of genes. These overlaps increase with $n$ and reach reasonable values ($\sim 0.5$) only when $n \sim 1000$. Note that the OPGLS of these datasets were obtained from training sets of a few hundred samples or less.

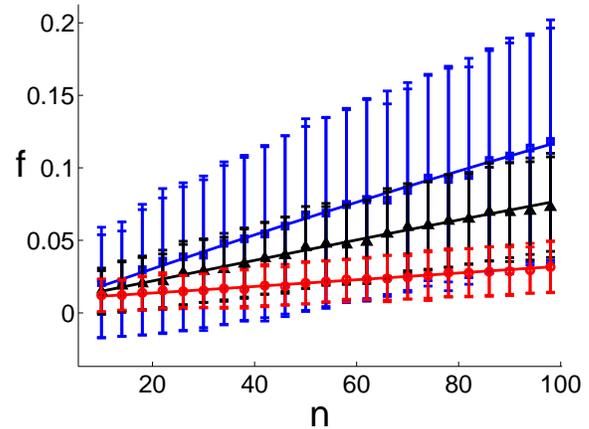

Figure 3: The mean overlap between the true and the observed OPGLs is plotted as a function of $n$. The three curves represent three distinct cancer types. Analytical results (solid lines) coincide with simulations for Hepatocellular Carcinoma [7] (blue squares), Lung Cancer [6] (black triangles), and pediatric ALL [9] (red circles). Errorbars representing st.d. values also show a close agreement between simulations and analytical results.

## 6 Discussion

We have provided an analytical framework for studying the robustness of ranking in the presence of noise. We explored the effect of noise level on the proximity of the observed ranking to the true one. We used two simple distinct similarity measures to quantify this proximity, each highlighting a different aspect of the obtained ranking. Other measures, as well as combinations of our proposed measures can be similarly approached, thus extending the potential range of applications. We successfully applied our method to study feature selection under noise, by investigating the instability of outcome predictive gene lists in several cancer types. We showed that the number of samples necessary to achieve reasonably reliable lists is at least an order of magnitude larger than the number of samples in currently available experiments. Simulation results supported our analytical predictions. The feature selection problems we studied included only ranking of individual genes. However, our method can be easily applied to more involved cases, where pairs, triples and multiples of variables are also ranked, simply by treating them as new objects. Interesting future directions include studying the asymptotic behavior of the solutions, and modifying the model to include correlations between the various noises. A different yet related problem is studying the overlap between two or more noisy rankings, obtained by different agents each of which views a different noisy version of the true values. Studying the overlaps between several noisy rankings may have application to rank aggregation problems [26]. This was studied for the case of two rankings in the context of outcome prediction in gene-expression cancer experiments [27].

**Acknowledgements**

This work was supported by grants from the Ridgefield Foundation, the Wolfson Foundation, and by European Community FP6 funding.